# A Comparative Study of Object Trackers for Infrared Flying Bird Tracking

Ying Huang[1], Hong Zheng[1], Haibin Ling[2], Erik Blasch[3], Hao Yang[4]

[1]School of Automation Science and Electrical Engineering, Beihang University, Beijing, China

[2]Department of Computer and Information Science, Temple University, Philadelphia, USA

[3]US Air Force Research Laboratory, Wright-Patterson Air Force Base, USA

[4]Airport Research Institute, China Academy of Civil Aviation Science and Technology, Beijing, China

*Abstract*

*Bird strikes present a huge risk for aircraft, especially since traditional airport bird surveillance is mainly dependent on inefficient human observation. Computer vision based technology has been proposed to automatically detect birds, determine bird flying trajectories, and predict aircraft takeoff delays. However, the characteristics of bird flight using imagery and the performance of existing methods applied to flying bird task are not well known. Therefore, we perform infrared flying bird tracking experiments using 12 state-of-the-art algorithms on a real BIRDSITE-IR dataset to obtain useful clues and recommend feature analysis. We also develop a Struck-scale method to demonstrate the effectiveness of multiple scale sampling adaption in handling the object of flying bird with varying shape and scale. The general analysis can be used to develop specialized bird tracking methods for airport safety, wildness and urban bird population studies.*

## 1. Introduction

The prevention of bird strike has always been the focus of guarantying aircraft flight safety. In the past few decades, flying bird surveillance relies mostly on human observations. However, the effectiveness of this fashion is limited in the poor visual conditions, such as at dawn and at dusk. Birds tend to be active during these times which requires extra vigilance. Many countries and aviation organizations desire automatic flying bird monitoring technology which can reduce or mitigate the risks of bird strikes to aircraft whether commercial, recreational, or military aircraft.

As a promising approach, computer vision technology has been proposed to monitor birds surrounding the airport. The basic principle of the idea is to use infrared and color cameras to scan entire airspace around the aerodrome, and automatically utilize visual techniques to detect and track flying birds from pictures, simultaneously transmit the tracking results to control tower and bird repelling equipment, and alerts and warnings to pilots affected.

In the field of computer vision, numerous powerful object tracking methods are developed to track pedestrians, vehicles or other common objects. For example, the algorithms used in these methods include sparse representation and particle filter [1], tracking based on boost detection [2], structured output tracking [3], and correlation filter based tracking [4], etc.

These algorithms obtained satisfying results in precision, operation speed, overcoming illumination variation, and occlusion.

To the best of our knowledge, a few of papers have concerned with flying bird tracking. Flying bird is a fast moving small object, it is hard to distinguish from the background in the infrared images, and its shape and scale are changing with time. For such a special object, how about the performance of existing tracking methods to track it, what are the strengths and weaknesses of each method on this problem, are not well understood by us. Thus we conduct a comprehensive experiment to study its characteristics and specificities. The experiment analysis also can inspire the community to continue to develop a dedicated approach for flying bird tracking, which not only can improve the safety of air transportation but can evaluate bird ecological populations in the wilderness and urban situations.

The paper is organized as follows: the flying bird surveillance system we built and collected bird dataset are presented in Section 2. Section 3 briefly reviews twelve state-of-the-art tracking methods and one modified enhanced structured output tracking method by ourselves. The experiment evaluation methodology is described in Section 4. Section 5 presents experiment results and analysis of the effect of various configuration settings to tracking performance. Finally, the work concludes with Section 6.

## 2. Experiment equipment and dataset

The flying bird surveillance system is equipped with a thermal and a visible light pan-tilt-zoom camera. The exterior structure of the BIRD Surveillance Infrared-Vis Tracking Exploitation (BIRDSITE) system is shown in Figure 1. The left circular window is infrared camera and the right part is visible light camera. The pan/tilt mechanism below provides accurate pointing control while giving full-space scanning.

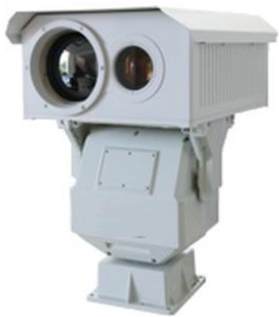

**Figure 1. Flying bird surveillance system**

For camera control, we developed a professional surveillance software to display the infrared and visible light images in real time, adapt the parameters of cameras, and control the pan/tilt camera angles. The software embedded automatic object detection, tracking and control modules to keep the flying bird in the center of camera's field of view (FOV). The images and all related information are saved to large scale database for further processing.

In the experiments, we use a domestic pigeon as the tracking target and utilize BIRDSITE system to collect flying bird images, obtained an infrared flying bird dataset, which we named BIRDSITE-IR Dataset. The dataset contains 17 video sequences, the frame rate of all sequences is 25 frames per second. To perform tracking evaluation, we annotated the position and scale of all targets in the sequences. It is noted that we annotate the center of bounding box at the body center of bird. Certainly bounding boxes exist more background region in order to include wing parts of bird. Generally, using the body center of bird as the position of object is more robust in evaluation studies. Furthermore, some bird repelling equipment, like laser and acoustic wave, would require its direction pointing to the body center of bird.

## 3. Evaluated object tracking methods

For conducting a comprehensive evaluation, we select 12 state-of-the-art online learning object tracking algorithms to perform flying bird tracking, which include L1APG[5], SET[6], CT[7], DFT[8], ASLA[9], CSK[10], SCM[11], LOT[12], Struck[3], HoughTrack[13], PixelTrack[14] and KCF[4]. In addition, we modified Struck method to enable it can adapt the variation of object scale in this paper. Online learning based tracking methods is now widely recognized since it exhibits the robustness to the change of object appearance, illumination variation, and occlusion. These methods were top ranked in the benchmarks [15][16][17][18] between 2013 and 2015.

In these mentioned methods, L1APG uses sparse representation to build object model, and utilizes particle filter to estimate object location. ASLA uses structured local sparse appearance model to improve tracking accuracy and handle occlusion. SCM builds a coordinate sparse model to describe a holistic and local object information for the abrupt variation of object appearance and drifting problems. KCF, SET, and CSK are based on correlation filter tracking framework, and CSK only use gray pixel value as feature, KCF adopts HOG feature and kernel function to promote tracking precision, SET adds an object scale estimation module after estimating object location. These three methods are very fast since correlation filter framework only calculate features one time in a wide search range, and utilizing fast fourier transformation to accelerate object position calculation. HoughTrack and PixelTrack are designed for non-rigid object tracking, which capitalize on random forest and pixel values based general Hough transformation models to calculate object location respectively, and then utilize GrabCut [19] and recursive Bayesian methods to perform segmentation for updating respective object models. LOT computes EMD (Earth Mover's Distance) and uses object templates or histograms to search the most probable candidate target. Struck chooses structured output SVM to estimate object location, and collects positive and negative samples around this location to update classifier online. The use of both object and background models enhances the discriminative power to the object.

Considering Struck method only samples in a constant scale, but the scale of flying bird often changes in time, we utilized a scale adaptation method from [20] to strengthen Struck method. In order to distinguish our method from the Struck algorithm, we denote the modified Struck algorithm as *Struck-scale*.

In the sampling part of Struck method, we denote the width and height of the bounding box in the previous frame as $w$ and $h$ respectively. The possible sizes of object in the current frame are as follows:

Table 1. The variation of object scale

| | | |
|---|---|---|
| $w' = w - \Delta w$, $h' = h - \Delta h$ | $w' = w - \Delta w$, $h' = h$ | $w' = w - \Delta w$, $h = h + \Delta h$ |
| $w' = w$, $h' = h - \Delta h$ | $w' = w$, $h' = h$ | $w' = w$, $h = h + \Delta h$ |

| $w' = w + \Delta w$, | $w' = w + \Delta w$, | $w = w + \Delta w$, |
| :---: | :---: | :---: |
| $h' = h - \Delta h$ | $h' = h$ | $h' = h + \Delta h$ |

We choose the value $\Delta w = 0.1w$ and $\Delta h = 0.1h$ in this paper. Sampling with different possible sizes provide the ability of scale adaptation in the tracking. In order to compare samples, we then scale all samples to the same size and measure their similarity.

## 4. Evaluation methodology

For the evaluation of tracking performance, we usually measure an averaged Euclidean distance between tracked results and actual object positions as the overall performance of one method on one sequence. However, the tracking results are random once tracking failure occurs. In this case, using an average error to describe the whole performance of one sequence is not accurate. We here adopt two more exact evaluation methodologies used in [15]. The first method, a *tracking precision* measure, is computed as the percentage of frames in the sequence such that the distance between tracked location and ground truth is within a given threshold. The second method, *tracking success* measure, is defined as the percentage of frames in which the overlap between tracked bounding box and ground truth bounding box exceeds a fixed threshold. The overlap of two bounding boxes $\tau$ is given by:

$$\tau = \frac{|r_t \cap r_a|}{|r_t \cup r_a|} \tag{1}$$

where $r_t$ is tracked bounding box, $r_a$ is annotated bounding box, ∩ and ∪ denote the intersection and union of two boxes respectively, and $|\cdot|$ represent the number of pixels in the given regions. In order to observe the tendency of tracking results, we select a threshold range of [1, 50] pixels for tracking precision and a threshold range of [0, 1] for tracking success, and calculate these two measures in each threshold value to generate corresponding plots. According to [4][15] and PASCAL evaluation criteria, we also report the scores of two performance measures at the threshold values of 20 pixels and 0.5 respectively.

## 5. Evaluation results

We first present the qualitative evaluation of 5 tracking methods for instance due to paper space limitations. Then we give the quantitative evaluation of 12 tracking methods. In addition, we compare the effect of variant configuration settings to tracking method. At last we analyze the effectiveness of enhanced Struck-scale method.

### 5.1. Qualitative evaluation

Figure 2 presents the tracking results of the first sequence, where we can observe the tracking precision of one method from each row and compare the results of multiple methods from each column. We can see that the variation of bird's shape and scale is dramatic in the flight and camera cannot to capture the transition period of flapping wings of bird in two adjacent frames. From the tracking results, we noticed that Struck and PixelTrack methods exist some tracking errors and bounding boxes cannot cover entire bird region. HoughTrack can adapt the change of bird's shape and scale, whereas KCF and LOT occur tracking lost.

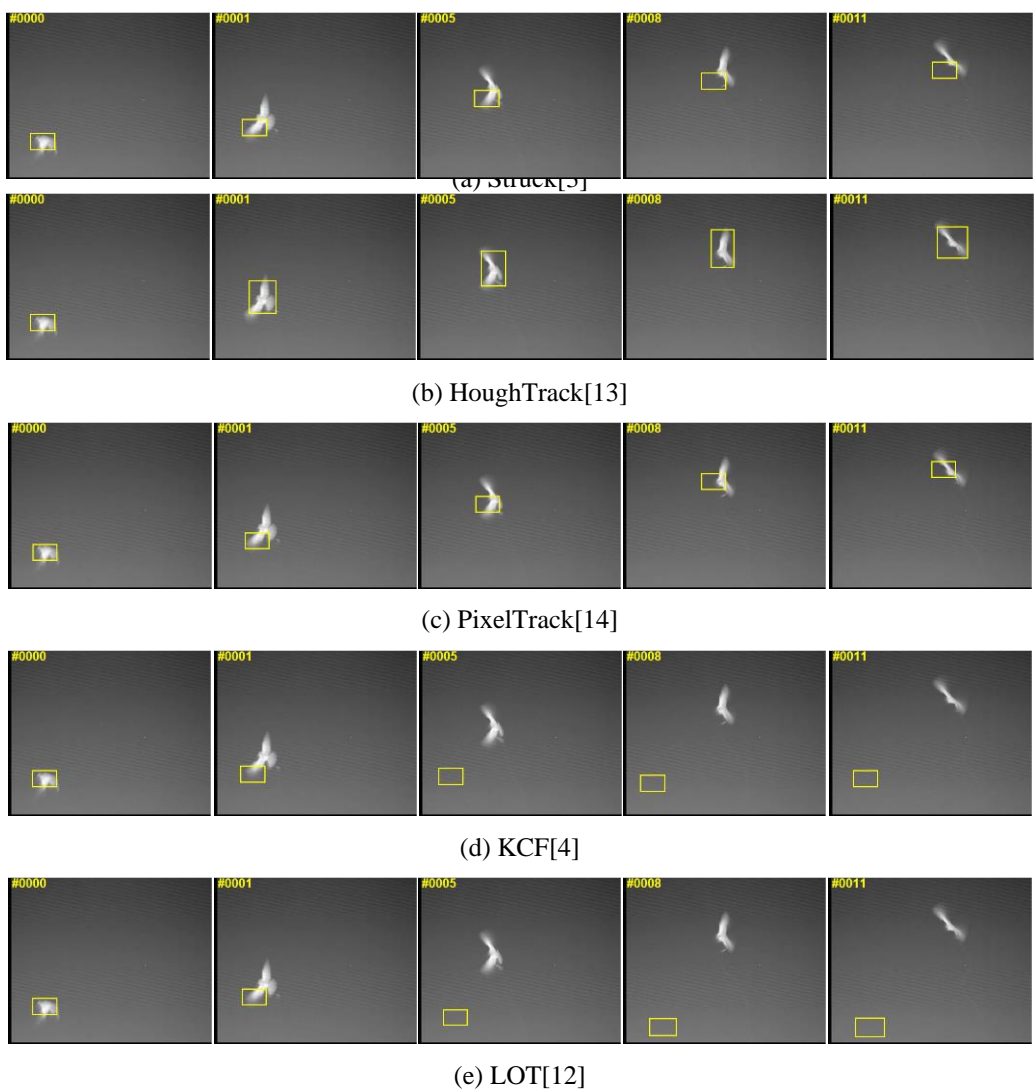

(a) Struck[3]

(b) HoughTrack[13]

(c) PixelTrack[14]

(d) KCF[4]

(e) LOT[12]

**Figure 2. The results of 5 tracking methods on sequence 1**

Figure 3 shows the tracking results of the second sequence, where there are some clouds and buildings. From the results, we can see that Struck method exists some tracking errors, and HoughTrack can track bird accurately, but PixelTrack, KCF and LOT methods lost target in the tracking.

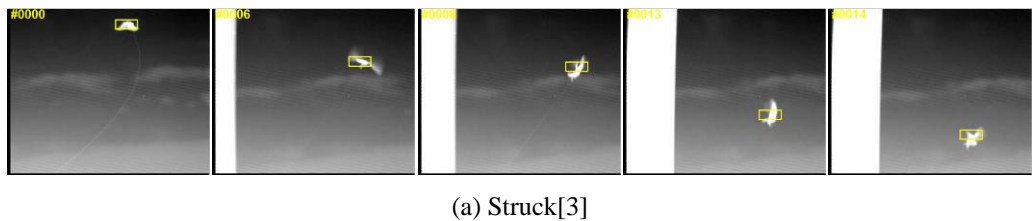

(a) Struck[3]

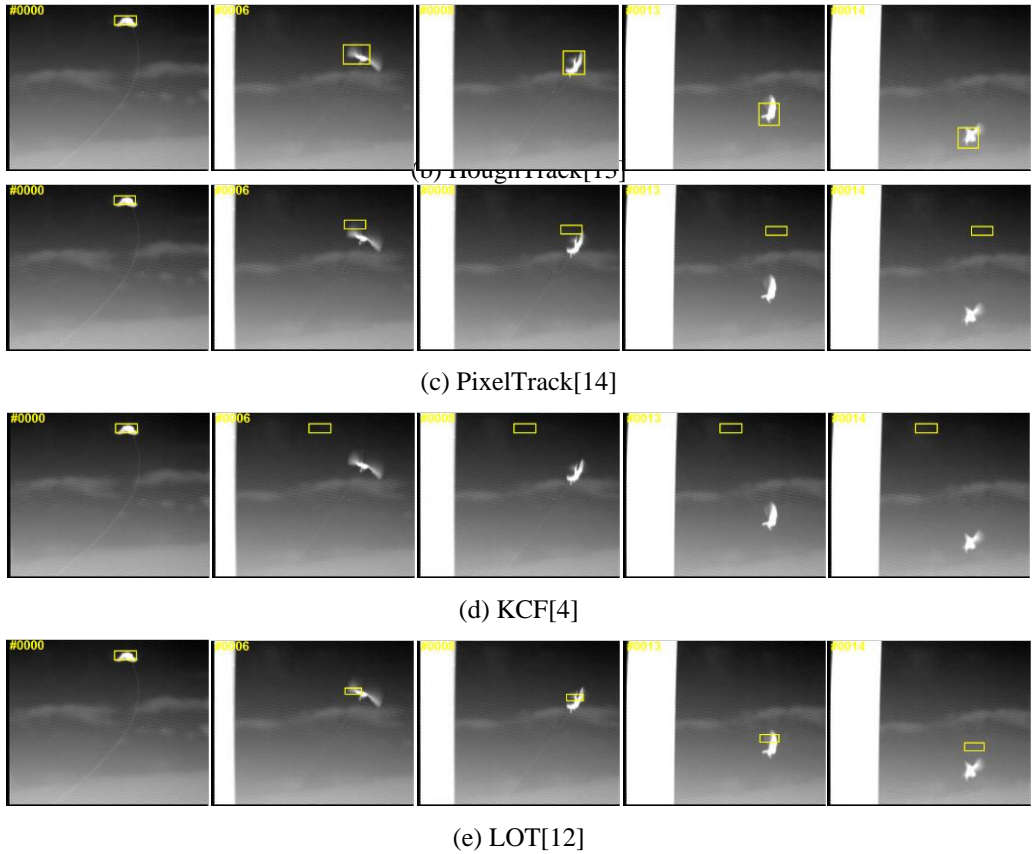

(b) HoughTrack[13]

(c) PixelTrack[14]

(d) KCF[4]

(e) LOT[12]

**Figure 3. The results of 5 tracking methods on sequence 2**

In Figure 4, there exist forest, building structure, ground, and clouds, and the scale of bird is small that about 10 pixels in height and width. Ellipses with red contour show magnified object region. The bird when it flying too low is easily be confused by forest and ground. From tracking results, we can find that Struck can track bird exactly in this complex situation, whereas the result region of HoughTrack far exceeds the scale of bird and includes substantial background objects, and PixelTrack, KCF, and LOT lost target in the tracking.

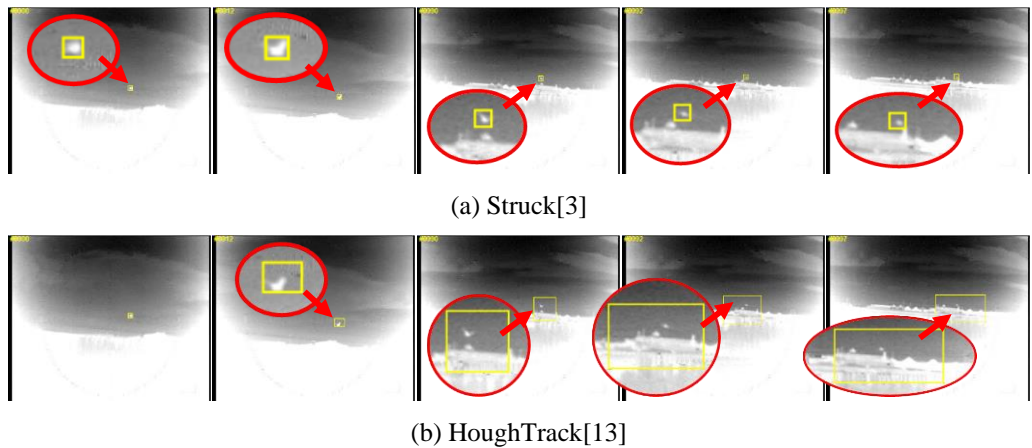

(a) Struck[3]

(b) HoughTrack[13]

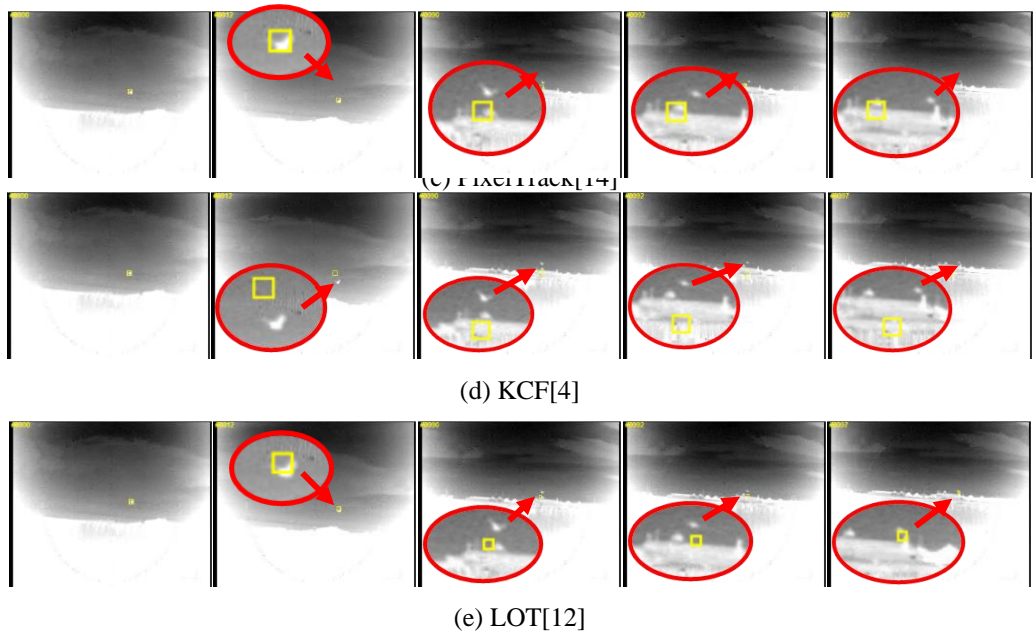

(e) LOT[12]

**Figure 4. The results of 5 tracking methods on sequence 3**

From above tracking results, we find that structured output SVM provides a powerful discrimination to Struck even in a cluttered situation, but it cannot adapt to the variation of object shape and scale and causes tracking errors. HoughTrack benefited from the use of segmentation so it can follow the change of object shape and scale, however, Hough forest is easily be disturbed by background when it estimates object location. The tracking failures of KCF and LOT illustrate that correlation filter framework used by both methods is sensitive to the object deformation. The object models in PixelTrack rely on pixel values, therefore causing its insufficient distinguishing ability in the complex infrared background.

We also should be noted that these online learning methods has a hypothesis that the change of object should be consecutive in an image sequence with a certain frame rate. It is to say, the object should not have a sudden variation between two adjacent frames such as being measured by a similarity metric. A tracking algorithm may be confused by object with large deformation and regard it as the background. In the flying bird tracking, once this assumption of similarity is not satisfied, it will lead to tracking error or failure.

**5.2. Quantitative evaluation**

In order to evaluate the overall performance of 12 tracking methods, we assess these methods on BIRDSITE dataset from two aspects of tracking precision and tracking success. Both precision and success plots are shown in Figure 5, where we can observe that as the threshold value gradually increased the tracking precision of each method also improved. The legend in precision plot reports the tracking precision score of each method at the threshold value of 20 pixels. The score of Struck (0.941) is almost 11% higher than the score of HoughTrack (0.847), which is mainly due to HoughTrack is easier distract than Struck when bird is flying near ground.

We also observed that while the threshold of overlap gradually raised, the number of frames whose overlapping with ground truth exceeds threshold value is decreased. The legend in success plot reports the success score of each method at a threshold value of 0.5, in which Struck and HoughTrack are similar, but better than other methods. From the quantitative evaluation of tracking precision and tracking success, the best method is Struck among these 12 tracking approaches.

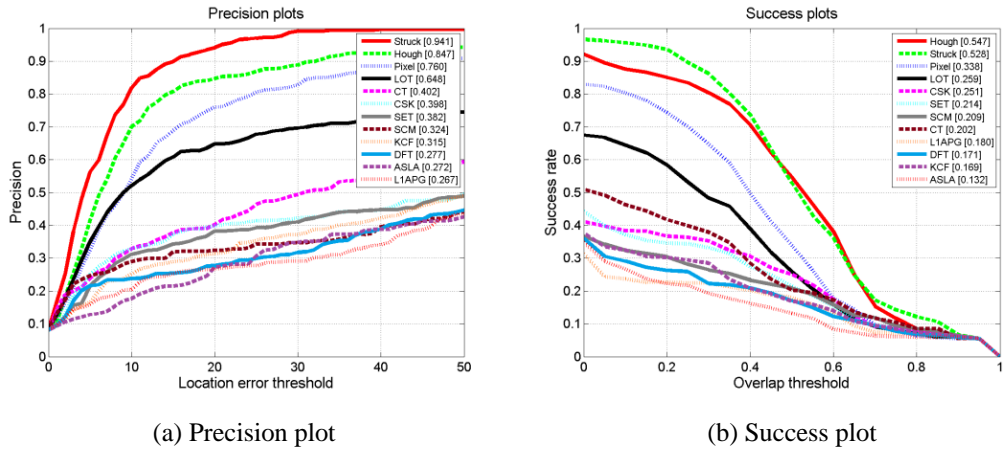

(a) Precision plot　　　　　　　　　　　　　(b) Success plot

**Figure 5. The plots of tracking precision and tracking success**

### 5.3. Feature configurations

From previous sections we analyzed the best tracking method in the flying bird tracking task. But we should noticed that a key factor that relates to tracking performance is building effective object appearance model [21]. In previous experiments Struck adopts Haar-like feature to describe object. Is there have other features more fit for flying bird representation and how about their tracking performance, are also worth to study.

Considering the limitation of infrared gray image and runtime speed, we select four easily computed common features: Haar-like feature, Histogram of oriented gradient (HOG) feature, gray-level histogram feature, and raw pixel feature. These features are often used in the object detection and tracking methods and display excellent performance and high computation efficiency. Because Struck uses kernel function to map data, we also choose four types of kernel function for analyzing their effect, which include Gaussian kernel with standard deviation $\sigma = 0.2$, Gaussian kernel with $\sigma = 0.1$, intersection kernel, and linear kernel. Four features and four kernel functions generate 16 configuration settings, which are used in Struck for performance analysis.

Figure 6 presents tracking precision and success plots with four types of kernel function. Each subplot is with one type of kernel function, where tracking precision and success curves of Haar, gray histogram, and raw features are close to each other, but the precision and success curves of HOG feature are significantly lower than other three features. This phenomena could be explained from each feature representation fashion. Haar, gray histogram, and raw features calculates the distribution of image gray intensity, whereas HOG feature counts the directions of pixel gradient, which is relevant to object shape. Therefore, the performance of HOG feature is limited in handling deformable object like flying bird.

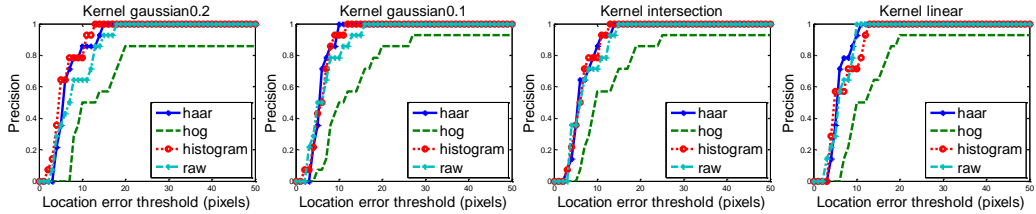

(a) Precision plots with four types of kernel functions

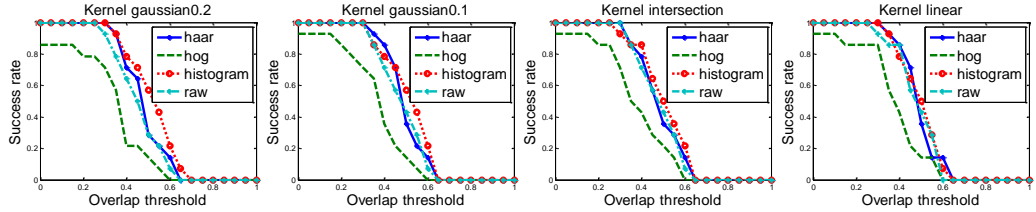

(b) Success plots with four types of kernel functions

**Figure 6. Precision and success plots with four types of kernel function**

Figure 7 shows the tracking precision and success plots with four types of feature. Each subplot is with one type of feature. We can find that the tracking precision and success curves of four variant kernel functions are quite similar, which illustrates that the effect of kernel function to tracking performance is less than feature selection.

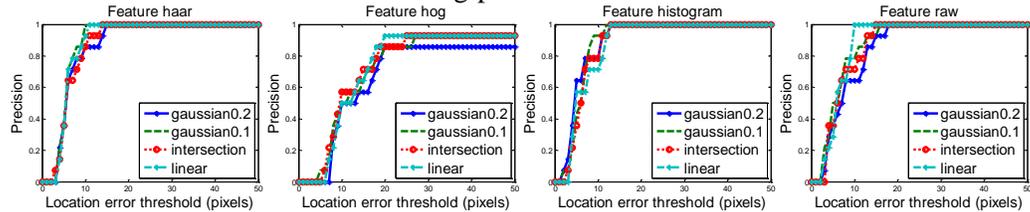

(a) Precision plots with four types of features

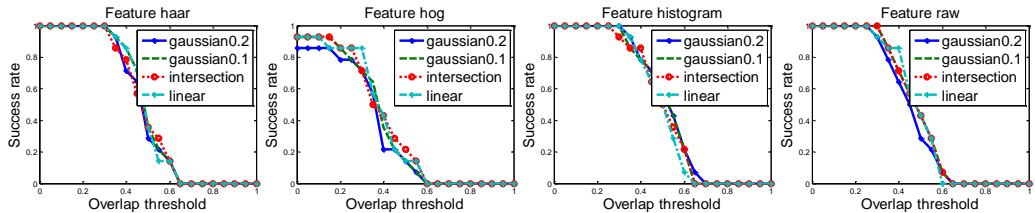

(b) Success plots with four types of features

**Figure 7. Precision and success plots with four types of feature**

### 5.4. Multi-scale analysis

Since the scale of flying bird is varying with time, we here utilize enhanced Struck-scale method to analyze the effectiveness of multi-scale adaptation strategy. Figure 8 and Figure 9 show the tracking precision and success comparisons of Struck and Struck-scale with 16 configuration settings. We can find that only Haar feature remains the same on both precision and success plots for two tracking methods, whereas other three features have an obvious decline on success plots. This indicates that the use of multi-scale adaptation strategy in the tracking method does not work in improving tracking performance while object's shape and scale change simultaneously. In addition, these comparisons also show the stability of Haar feature in object representation.

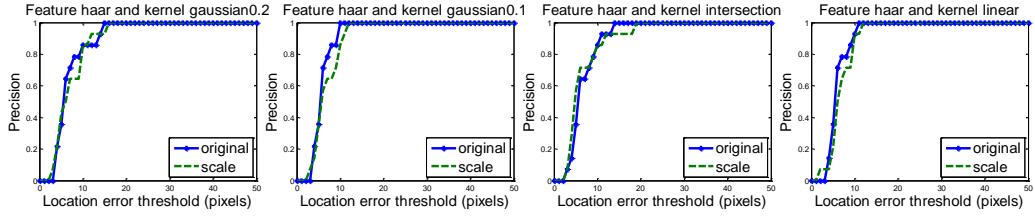

(a) Precision plots of Haar feature with four variant kernel functions

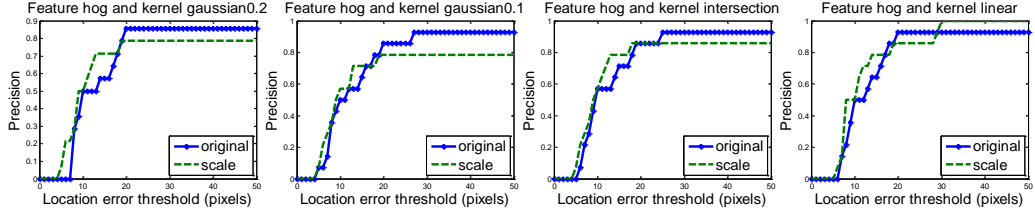

(b) Precision plots of HOG feature with four variant kernel functions

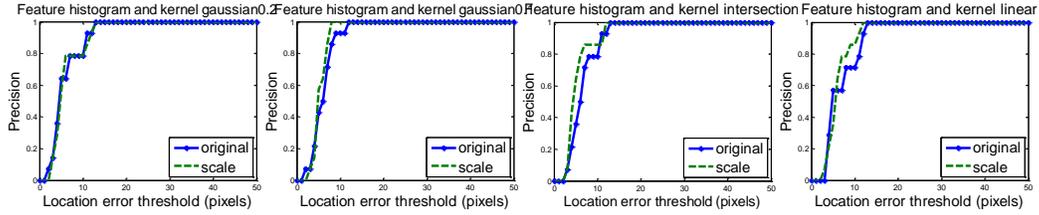

(c) Precision plots of histogram feature with four variant kernel functions

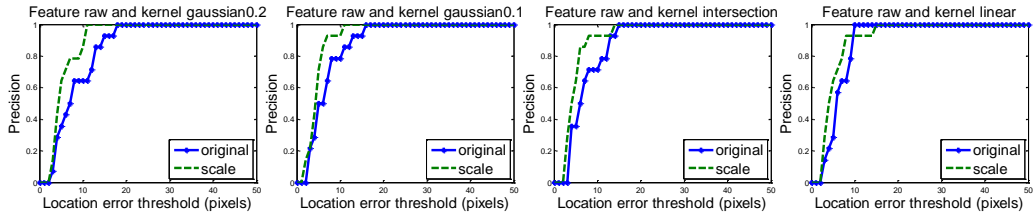

(d) Precision plots of raw feature with four variant kernel functions

**Figure 8. Tracking precision comparison of Struck and Struck-scale methods**

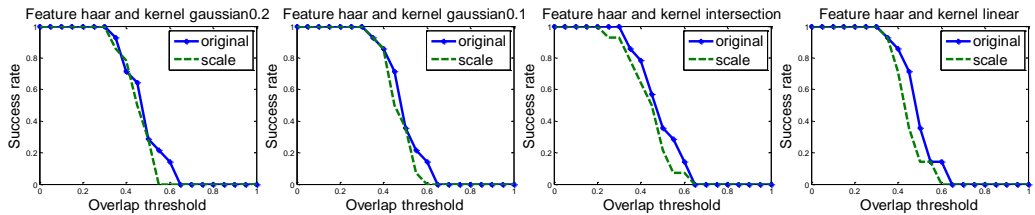

(a) Success plots of Haar feature with four variant kernel functions

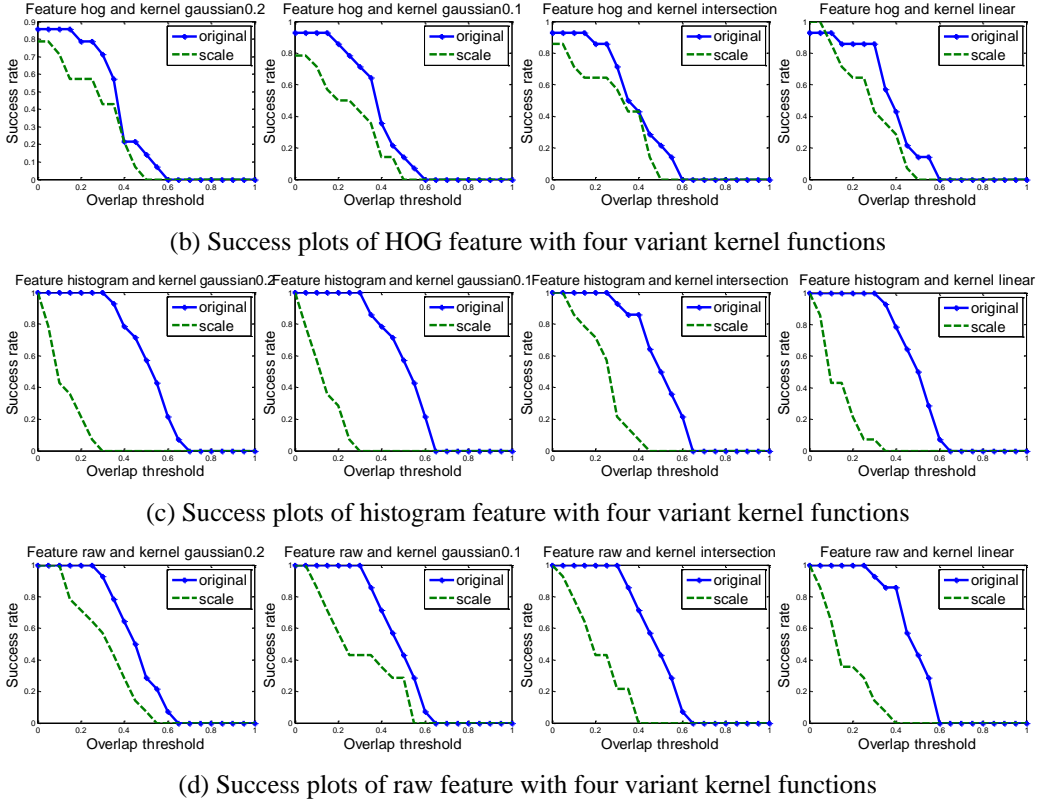

(b) Success plots of HOG feature with four variant kernel functions

(c) Success plots of histogram feature with four variant kernel functions

(d) Success plots of raw feature with four variant kernel functions

**Figure 9. Tracking success comparison of Struck and Struck-scale methods**

## 6. Conclusions

This paper evaluates the performance of 12 state-of-the-art tracking methods on infrared flying bird tracking task through both qualitative and quantitative comparisons, and analyzed the effect of variant features and kernel functions to tracking performance. Finally, we use modified Struck-scale method to demonstrate the effectiveness of multi-scale adaptation strategy. According to the experiment results and observation, we have the following conclusions:

1) The shape and scale of bird can change drastically in the flight, which is the most important challenge to existing similarity metric based tracking methods.

2) Among 12 tracking algorithms, Struck method shows the best performance in the infrared flying bird tracking, and the second best method is HoughTrack. Struck has a great object discriminative power, but it cannot adapt the scale of bounding box and causes error in tracking deformable object. HoughTrack can adapt to the change of bird shape and scale, however, it is easily be disturbed by background and results in tracking failure.

3) The use of segmentation which enables HoughTrack can adapt to object deformation is an enlightening clue for developing a special flying bird tracking method. Compared with multiple scale sampling, segmentation has the advantages of inexpensive computation cost and extracting precise object region.

4) Gray histogram and raw features can describe the object which has some shape variation. The performance of Haar-like feature is steady in representing object with variant shapes and scales. However, these three features are not sufficiently robust to calculate flying bird like deformable object. HOG feature has a deteriorated property when it applies to flying bird tracking.

5) For tracking flying bird like deformable object, feature representation should have a complete shape invariance property, otherwise multi-scale sampling would not improve tracking performance.

6) Feature representation plays a more important role than kernel function for tracking performance.